\documentclass[runningheads]{llncs}
\usepackage{graphicx, booktabs, paralist}
\usepackage[table,xcdraw]{xcolor}

\usepackage{booktabs}
\usepackage{graphicx}
\usepackage[table,xcdraw]{xcolor}
\usepackage{amsmath}
\usepackage{amsfonts}
\begin{document}
\title{Principled Ultrasound Data Augmentation for Classification of Standard Planes}
%
%
\author{Lok Hin Lee, Yuan Gao and J. Alison Noble}
\authorrunning{L.H.L. et al.}
%
\institute{Department of Engineering Science, University of Oxford, UK}
\maketitle              
\begin{abstract}

Deep learning models with large learning capacities often overfit to medical imaging datasets. This is because training sets are often relatively small due to the significant time and financial costs incurred in medical data acquisition and labelling. 
Data augmentation is therefore often used to expand the availability of training data and to increase generalization. 
However, augmentation strategies are often chosen on an ad-hoc basis without justification. 
In this paper, we present an augmentation policy search method with the goal of improving model classification performance. 
We include in the augmentation policy search additional transformations that are often used in medical image analysis and evaluate their performance. 
In addition, we extend the augmentation policy search to include non-linear mixed-example data augmentation strategies.
Using these learned policies, we show that principled data augmentation for medical image model training can lead to significant improvements in ultrasound standard plane detection, with an an average F1-score improvement of 7.0\% overall over naive data augmentation strategies in ultrasound fetal standard plane classification. 
We find that the learned representations of ultrasound images are better clustered and defined with optimized data augmentation.

\end{abstract}
\section{Introduction}

The benefits of data augmentation for training deep learning models are well documented in a variety of tasks, including image recognition \cite{luke,ce,ryo} and regression problems \cite{gz,ho}.
Data augmentation acts to artificially increase the size and variance of a given training dataset by adding transformed copies of the training examples.
This is particularly evident in the context of medical imaging, where data augmentation is used to combat class imbalance \cite{ze}, increase model generalization \cite{mf,dj}, and expand training data \cite{zab,df}.
This is usually done with transformations to the input image that are determined based on expert knowledge and cannot be easily transferred to other problems and domains. 
In ultrasound, this usually manifests as data augmentation strategies consisting of small rotations, translations and scalings \cite{azhp}. 
However, whilst it is appealing to base augmentation strategies on ``expected" variations in input image presentation, recent work has found that other augmentation strategies that generate ``unrealistic looking" training images \cite{slik,mixaug} have led to improvements in generalization capability.
There has therefore been great interest in developing data augmentation strategies to automatically generate transformations to images and labels that would lead to the greatest performance increase in a neural network model. 
In this paper, inspired by the RandAugment \cite{ebj} augmentation search policy, we automatically look for augmentation policies that outperform standard augmentation strategies in ultrasound imaging based on prior knowledge and extend our algorithm to include mixed-example data augmentation \cite{mixaug} in the base policy search. 
We evaluate the proposed method on second-trimester fetal ultrasound plane detection, and find that a randomly initialized network with our augmentation policy achieves performance competitive with methods that require external labelled data for network pre-training and self-supervised methods. We also evaluate our method on a fine-tuning a pre-trained model, and find that using an optimized augmentation policy during training improves final performance.

\subsubsection{Contributions:}
Our contributions are three fold: 1) We investigate the use of an augmentation search policy with hyperparameters that does not need expensive reinforcement learning policies and can be tuned with simple grid search; 2) We extend this augmentation search policy to combinations that include mixed-example based data augmentation and include common medical imaging transformations; 3) We explain the performance of optimal augmentation strategies by looking at their affinity, diversities and effect on final model performance.

\subsubsection{Related Work}
Medical image datasets are difficult and expensive to acquire. 
There has therefore been previous work that seeks to artificially expand the breadth of training data available in medical image classification \cite{mf,zhf}, segmentation \cite{nal,ze} and regression \cite{df}. 

Original Data Manipulation:
Zeshan et al. \cite{zhf} evaluate the performance of eight different affine and pixel level transformations by training eight different CNNs for predicting mammography masses and find that ensembling the trained models improves the classification performance significantly. 
Nalepa et al. \cite{nal} elastically deform brain MRI scans using diffeomorphic mappings and find that tumour segmentation is improved.
However, in the above works, specific augmentations and parameters are selected arbitrarily and are task and modality dependent. In contrast, we propose an automated augmentation policy search method that can out perform conventional medical imaging augmentation baselines. 

Artificial Data Generation:
Florian et al. \cite{df} generates new training samples in by linearly combining existing training examples in regression. Models trained to estimate the volume of white matter hyperintensities had performance comparable to networks trained with larger datasets.
Zach et al.\cite{ze} also linearly combine training examples and target labels linearly inspired by mix-up \cite{hzmc} but focus on pairing classes with high and low incidence together, which was found to be beneficial for tasks with high class imbalance such as in brain tumor segmentation. Maayan et al. \cite{mf} train a conditional generative adversarial network (cGAN) to generate different types of liver lesions and use the synthesized samples to train a classification network. Dakai et al. \cite{dj} use a cGAN to synthesize 3D lung nodules of different sizes and appearances at multiple locations of a lung CT scan. These generated samples were then used to finetune a pretrained lung nodule segmentation network that improved segmentation of small peripheral nodules. However, cGAN-based methods are difficult to train and have significant computational costs during augmentation.

Automated Augmentation Policy Search:
There are augmentation policy search methods in the natural image analysis \cite{edcb,dhel,slik} that learn a series of transformations which are parameterized by their magnitudes and probability. However, these searches are expensive, and cannot be run on the full training dataset as the hyperparameter search for each transformation require significant computational resources. RandAugment (RA) \cite{ebj} finds that transformations can have a shared magnitude and probability of application and achieve similar performance, without expensive reinforcement learning. However, RA is limited to single-image transformations. We therefore explore the use of an extended RA policy search with additional transformations that are more specific to medical imaging, and expand its capabilities to include mixed-example image examples to include artificial data in model training. 
%

\section{Methods}
\begin{figure}[h]
  \includegraphics[width=\textwidth]{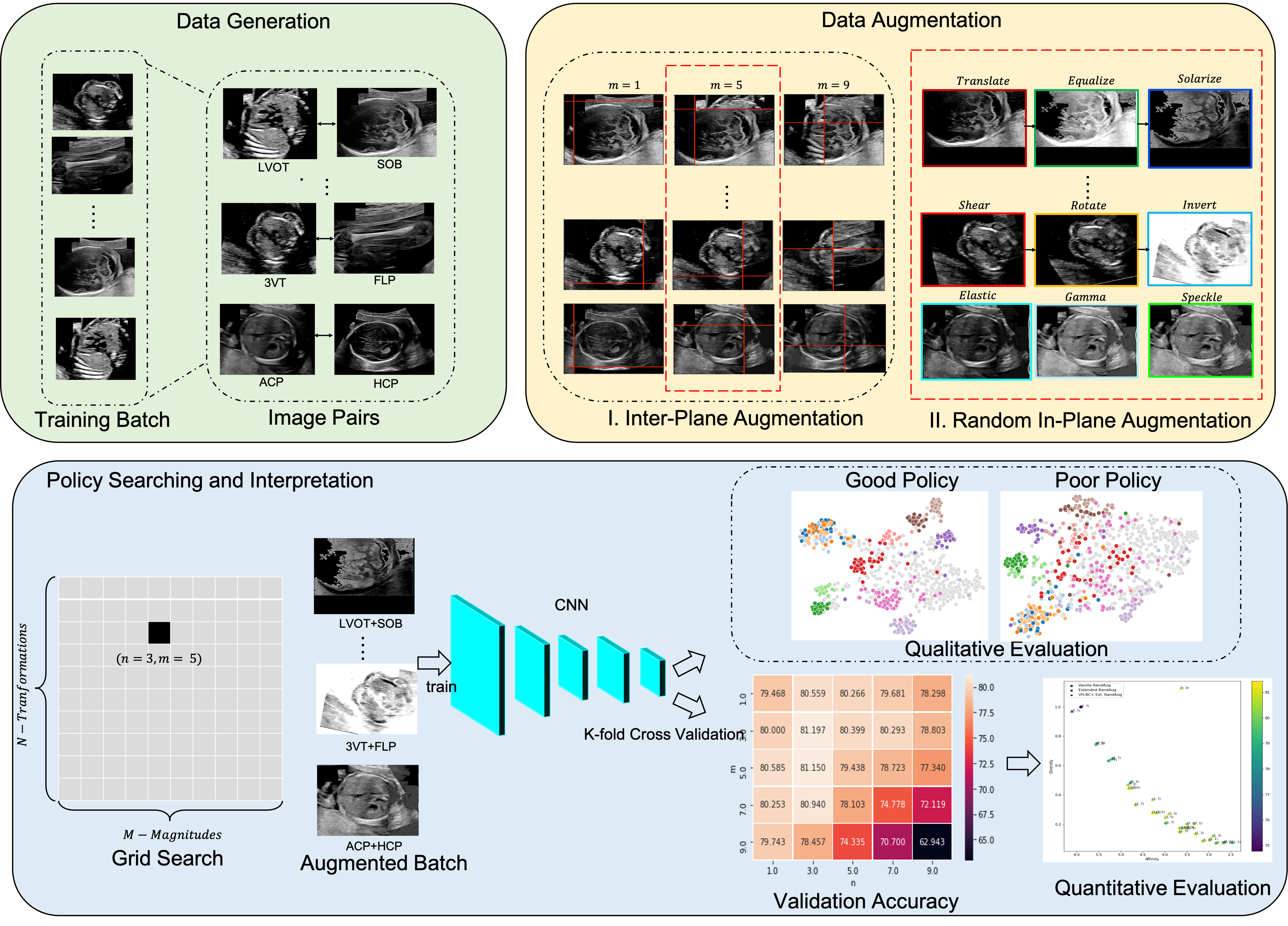}
  \caption{Overview of our proposed learning framework.}
  \label{pics:arch}
  \vspace{-.4cm}
\end{figure}
In this section we describe our proposed framework for augmentation policy search, depicted in Figure \ref{pics:arch} which consists of three key procedures i) data generation, ii) data augmentation, iii) policy searching and interpretation.

\subsubsection{Mixed-Example Data Augmentation}
\begin{figure}
  \includegraphics[width=\textwidth]{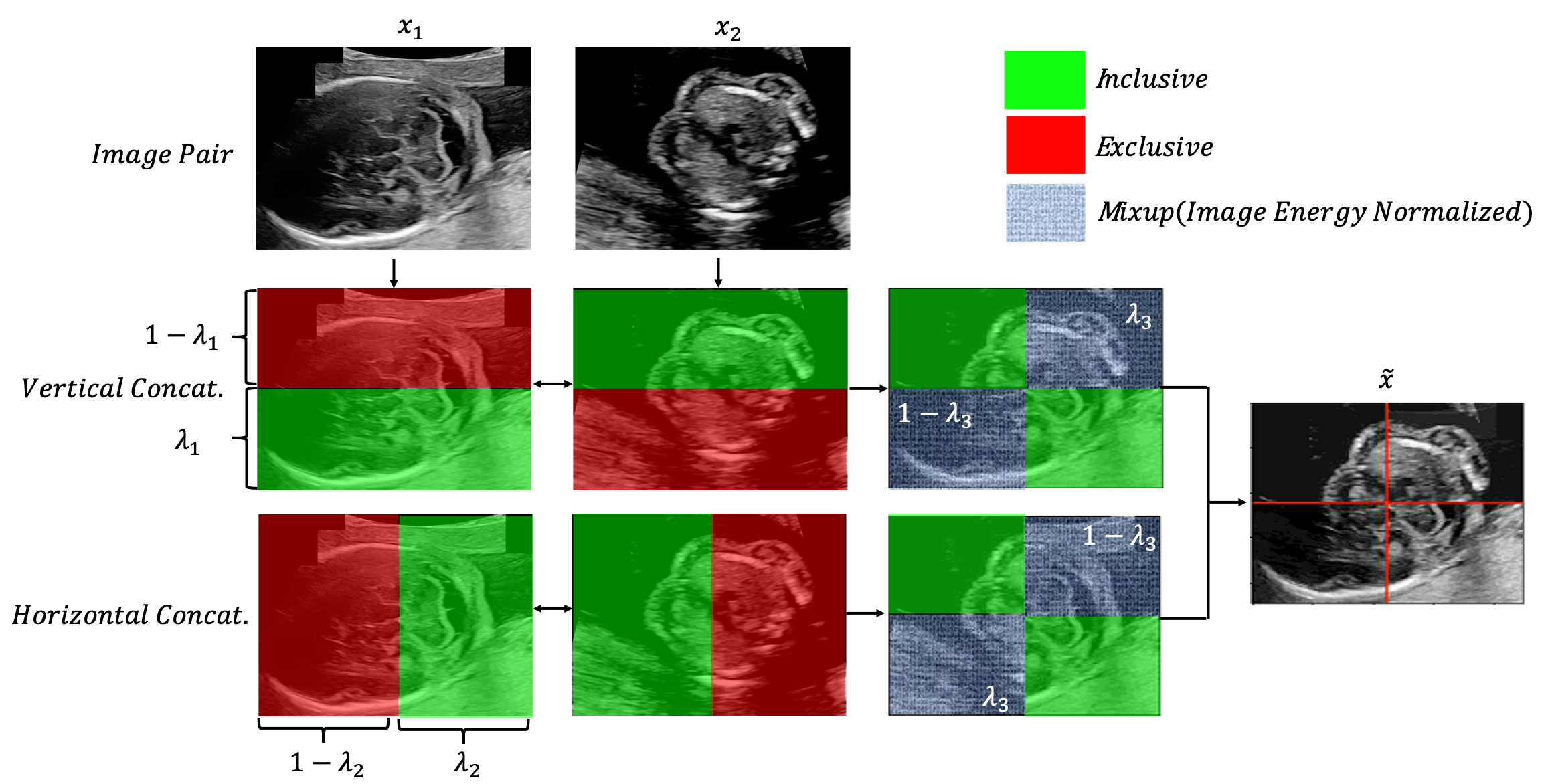}
  \caption{The procedure for non-linear mixed-example data augmentation using an image pair and the final artificial mixed-example image.}
  \label{pics:mixup}
  \vspace{-.4cm}
\end{figure}
The original dataset $D = \{(X_i, Y_i)\}$ consists of a series of $i$ ultrasound frames $X$ and their associated classes $Y$. We first generate a paired dataset $D_{paired} = \{(x_1, x_2)_\frac{i}{2}, (y_1, y_2)_\frac{i}{2}\}$ by pairing examples from different classes. 
Examples of artificial data are then generated using non-linear methods \cite{CMJ}, which are found to be more effective than linear intensity averaging (mix-up)\cite{hzmc}. As illustrated in Figure \ref{pics:mixup}, instead of pixel-wise averaging, the bottom $\lambda_{1}$ fraction of image $x_{1}$ is vertically concatenated with the top $1-\lambda_{1}$ fraction of image $x_{2}$. Similarly, the right $\lambda_{2}$ fraction of image $x_{1}$ is horizontally concatenated with the left $1-\lambda_{2}$ fraction of image $x_{2}$. After the concatenations, the resulted images are combined to produce an image $\tilde{x}$ in which the top-left is from $x_{1}$, the bottom right is from $x_{2}$, and the top-right and bottom-left are mixed between the two. Moreover, instead of linear pixel averaging, we treat each image as a zero-mean waveform and normalize mixing coefficients with image intensity energies \cite{tyy}. Formally, given initial images $x_{1,2}$ with image intensity means and standard deviations of $\mu_{1,2}$ and $\sigma_{1,2}$, the generated artificial mixed-example image $\tilde{x}$ is:
\[   
\tilde{x} = 
     \begin{cases}
       x_{1}(i,j)-\mu_{1} &\quad\text{if} \; i \le \lambda_{1}H \; \text{and}\; j\le\lambda_{2}W\\
       \frac{c}{\phi }[x_{1}(i,j)-\mu_{1}]+\frac{1-c}{\phi}[x_{2}(i,j)-\mu_{2}] &\quad\text{if} \; i \le \lambda_{1}H \; \text{and}\; j>\lambda_{2}W\\
       \frac{1-c}{\phi }[x_{1}(i,j)-\mu_{1}]+\frac{c}{\phi}[x_{2}(i,j)-\mu_{2}] &\quad\text{if} \; i > \lambda_{1}H \; \text{and}\; j\le\lambda_{2}W\\
       x_{2}(i,j)-\mu_{2} &\quad\text{if} \; i > \lambda_{1}H \; \text{and}\; j>\lambda_{2}W\\
     \end{cases}
\]
where $c$ is the mixing coefficient $(1+\frac{\sigma_{1}}{\sigma_{2}}\cdot \frac{1-\lambda_{3}}{\lambda_{3}})^{-1}$ and $\phi$ is the normalization term defined as $\sqrt{c^{2}+(1-c)^{2}}$. The row index and column index is represented by $i, j$ and the height and width of the images are represented by $H, W$.

We sample $\lambda_{1,2,3} \sim Beta(m/10,m/10)$ where $m$ is a learnt hyperparameter varied from 0-10. As m approaches 10, $\lambda$ values are more uniformly distributed across 0-1 and artificial images are more interpolated. The ground truth label after interpolation is determined by the mixing coefficients and can be calculated with: \[\tilde{y}=(\lambda_{3}\lambda_{1}+(1-\lambda_{3})\lambda_{2})y_{1}+(\lambda_{3}(1-\lambda_{1})+(1-\lambda_{3})(1-\lambda_{2}))y_{2}\]

\subsubsection{Original Data Augmentation}
Augmentation transformations are then applied to the mixed images. Inspired by \cite{ebj}, we do not learn specific magnitudes and probabilities of applying each transformation in a given transformation list. Each augmentation policy is instead defined only by $n$, which is the number of transformations from the list an image undergoes, and $m$, which is the magnitude distortion of each transformation. Note that $m$ is a shared hyperparameter with the mixed-example augmentation process. 
\begin{figure}
\centering
  \includegraphics[width=0.8\textwidth]{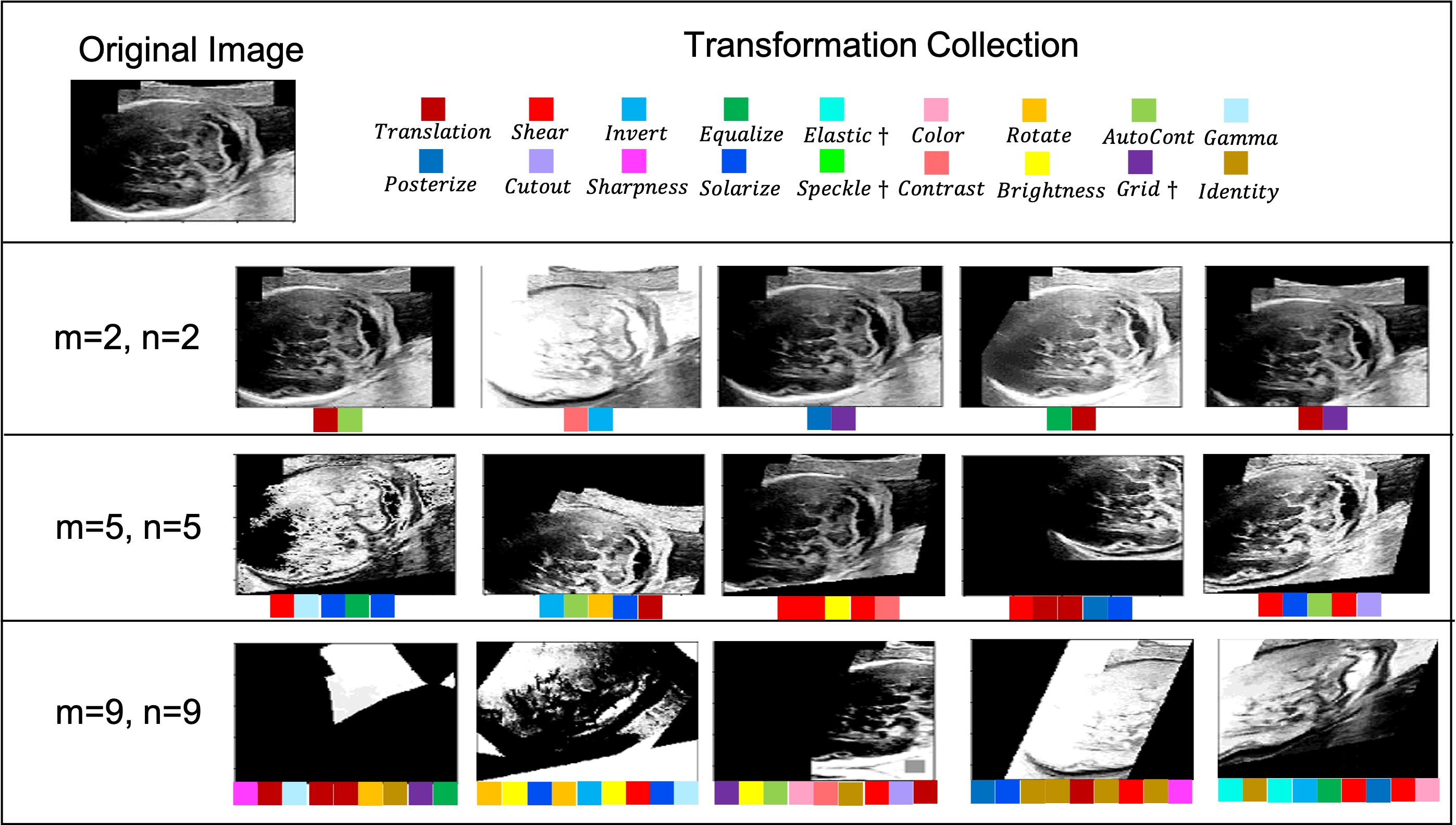}
  \caption{Examples of how each transformation and augmentation policy affect input images. Each color represents a transformation, and augmented image is transformed using a number of transformations (n) at a magnitude of (m). $\dagger$: our additional transformations.}
  \label{pics:ranaug}
  \vspace{-.4cm}
\end{figure}
We investigate the inclusion in the transformation list transformations commonly used in ultrasound image analysis augmentation: i) grid distortions and elastic transformation \cite{avi} and ii) speckle noise \cite{blsa}. The transformation list then totals 18 transformations, examples of which can be seen in Figure \ref{pics:ranaug}. 

\subsubsection{Optimization}
We define $f$ and $\theta$ as a convolutional neural network (CNN) and its parameters. As depicted in Figure \ref{pics:arch}, we train a CNN with the augmented mini-batch data $\tilde{x}^{i}$ and obtain the predicted output class scores $f_{\theta}(\tilde{x}^{i})$. These are converted into class probabilities $p(\tilde{x}^{i})$ with the softmax function. The KL-divergence between $f_{\theta}(\tilde{x}^{i})$ and $\tilde{y}^{i}$ is then minimized with back-propagation and stochastic gradient descent
\[L=\frac{1}{B}D_{KL}(\tilde{y}^{i}\parallel p(\tilde{x}^{i}))=\frac{1}{B}\sum_{i=1}^{B}\sum_{j=1}^{C}\tilde{y}_{j}^{i}log\frac{\tilde{y}_{j}^{i}}{\left \{ p(\tilde{x}^{i}) \right \}_{j}}\]
where $B$ is the batch size, $C$ is the number of classes and $L$ is the loss. 

Due to the limited search space, the hyperparameters $n$ and $m$ that produce the optimum classification performance can then be found using grid search as seen in Figure \ref{pics:arch}. The best performing $m, n$ tuple is then used during final model evaluation.
 
\subsubsection{Quantifying Augmentation Effects}
Next, we investigate how augmentation improves model generalization and quantify how different augmentation policies affect augmented data distributions and model performance. We adopt a two dimensional metric - affinity and diversity \cite{rsjj} to do this. Affinity quantifies the distribution shift of augmented data with respect to the unaugmented distribution captured by a baseline model; the diversity quantifies complexity of the augmented data. Given training and validation datasets, $D_{t}$ and $D_{v}$, drawn from the original dataset $D$, we can generate an augmented validation dataset $D(m, n)_{v}^{'}$ derived from $D_{v}$ using $m, n$ as hyperparameters for the augmentation policy. The affinity $A$ for this augmentation policy is then:
\[A=\mathbb{E}[L({D_{v}^{'}})]-\mathbb{E}[L(D_{v})]\]
where $\mathbb{E}[L(D)]$ represents the expected value of the loss computed on the dataset $D$ loss of a model trained on $D_{t}$. 

The diversity, $D$, of the augmentation policy $a$ is computed on the augmented training dataset $D_{t}^{'}$ with respect to the expected final training loss, $L_{t}$, as: 
\[D=\mathbb{E}[L(D_{t}^{'})]\]
Intuitively, the greater the difference in loss between an augmented validation dataset and the original dataset on a model trained with unaugmented data, the greater the distribution shift of the augmented validation dataset. Similarly, the greater the final training loss of a model on augmented data, the more complexity and variation there is in the final augmented dataset.

\section{Experiments and Results}
We use a clinically acquired dataset consisting of ultrasound second-trimester fetal examinations. A GE Voluson E8 scanner was used for ultrasound image acquisition. For comparison with previous work \cite{RichardIPMI,jbi}, fetal ultrasound images were labelled into 14 categories. Four cardiac view classes (4CH, 3VV, LVOT, RVOT) corresponding to the four chamber view, three vessel view, left and right ventricle tracts respectively; the brain transcerebellar and transventricular views (TC, TV); two fetal spine sagittal and coronal views (SpineSag, SpineCor); the kidney, femur, abdominal circumference standard planes, profile view planes and background images. The standard planes from 135 routine ultrasound clinical scans were labelled, and 1129 standard plane frames were extracted. A further 1127 background images were also extracted and three-fold cross validation was used to verify the performance of our network.

\subsubsection{Network Implementation}
 The performance of the SE-ResNeXt-50 \cite{SEResnet} backbone is well validated on natural images and therefore used. Networks were trained with an SGD optimizer with learning rate of $10^{-3}$, a momentum of 0.9 and a weight decay of $10^{-4}$. Networks were trained for a minimum of 150 epochs, and training was halted if there was 20 continuous epochs without improvement in validation accuracy. Models were implemented with PyTorch and trained on a NVIDIA GTX 1080 Ti. Random horizontal and vertical flipping were used in all RA policies as a baseline augmentation. Models were trained with a batch size of 50. We evaluated the performance of networks trained with augmentation policies with values of $m, n$ where $m, n = \{1, 3, 5, 7, 9\}$ and used a simple grid search for augmentation strategies to find optimal $m, n$ values. 

\subsubsection{Results on CNNs with Random Initialization}
\begin{table}\centering
\caption{Results for standard plane detection (mean $\pm$ std $\%$). The best performing augmentation strategies are marked in \textbf{bold} for each metric.}
\resizebox{\textwidth}{!}{%
\setlength{\tabcolsep}{8pt}
\begin{tabular}{@{}l|cccc|cccc@{}}
{\color[HTML]{000000} }          & \multicolumn{4}{l|}{{\color[HTML]{000000} Random Initialization}}                                                                                                                                                                            & \multicolumn{4}{l|}{{\color[HTML]{000000} Initialized with external data}}                                                                                                                                                                                                                                                                                                                     \\ \midrule
{\color[HTML]{000000} }          & {\color[HTML]{000000} No Aug.}      & {\color[HTML]{000000} SN Pol.}      & {\color[HTML]{000000} \begin{tabular}[c]{@{}c@{}}RA\\ \cite{ebj}\end{tabular}} & {\color[HTML]{000000} \begin{tabular}[c]{@{}c@{}}Mix. RA\\ (ours)\end{tabular}} & {\color[HTML]{000000} \begin{tabular}[c]{@{}c@{}}Siam. Init.\\ \cite{jbi}\end{tabular}} & {\color[HTML]{000000} \begin{tabular}[c]{@{}c@{}}Saliency\\ \cite{RichardIPMI}\end{tabular}} & {\color[HTML]{000000} \begin{tabular}[c]{@{}c@{}}SonoNet\\ \cite{RichardIPMI,Sononet}\end{tabular}} & {\color[HTML]{000000} \begin{tabular}[c]{@{}c@{}}SonoNet + Mix.RA\\ (ours)\end{tabular}} \\ \midrule
{\color[HTML]{000000} Precision} & {\color[HTML]{000000} 56.5$\pm$1.2} & {\color[HTML]{000000} 70.4$\pm$2.3} & {\color[HTML]{000000} 74.7$\pm$1.8}                                            & {\color[HTML]{000000} \textbf{75.1$\pm$1.8}}                                    & {\color[HTML]{000000} 75.8$\pm$1.9}                                                            & {\color[HTML]{000000} 79.5$\pm$1.7}                                                          & {\color[HTML]{000000} 82.3$\pm$1.3}                                                                 & {\color[HTML]{000000} \textbf{86.3$\pm$1.3}}                                                                \\
{\color[HTML]{000000} Recall}    & {\color[HTML]{000000} 55.1$\pm$1.2} & {\color[HTML]{000000} 64.9$\pm$1.6} & {\color[HTML]{000000} 72.2$\pm$2.3}                                            & {\color[HTML]{000000} \textbf{73.4$\pm$1.9}}                                    & {\color[HTML]{000000} 76.4$\pm$2.7}                                                            & {\color[HTML]{000000} 75.1$\pm$3.4}                                                          & {\color[HTML]{000000} \textbf{87.3$\pm$1.1}}                                                                 & {\color[HTML]{000000} 85.1$\pm$1.3}                                                                \\
{\color[HTML]{000000} F1-Score}  & {\color[HTML]{000000} 55.4$\pm$1.2} & {\color[HTML]{000000} 67.0$\pm$1.3} & {\color[HTML]{000000} 72.8$\pm$1.8}                                            & {\color[HTML]{000000} \textbf{74.0$\pm$1.8}}                                    & {\color[HTML]{000000} 75.7$\pm$2.0}                                                            & {\color[HTML]{000000} 76.6$\pm$2.6}                                                          & {\color[HTML]{000000} 84.5$\pm$0.9}                                                                 & {\color[HTML]{000000} \textbf{85.4$\pm$1.5}}                                                                \\ \bottomrule
\end{tabular}%
}
\label{tab:results}
\end{table}

The effectiveness of our mixed-example augmentation search policy algorithm (\textit{Mix. RA}) on SE-ResNeXt-50 models that are randomly initialized is compared with models trained with the baseline RandAugment (\textit{RA}) augmentation search policy; a commonly used augmentation strategy (\textit{SN Pol.}) in line with that found in \cite{RichardIPMI}, where images are augmented with random horizontal flipping, rotation $\pm10^{\circ}$, aspect ratio changes $\pm10\%$, cropping and changing brightness $\pm25\%$ and image cropping $95-100\%$; and no augmentation (\textit{No. Aug.}). 

\begin{figure}

\centering
\includegraphics[width=.49\textwidth]{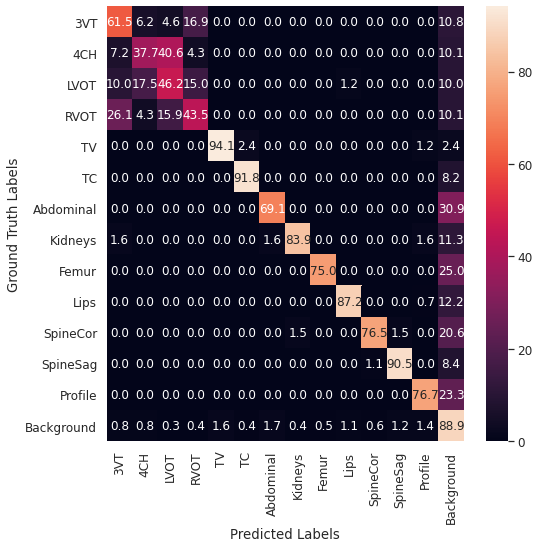}
\includegraphics[width=.49\textwidth]{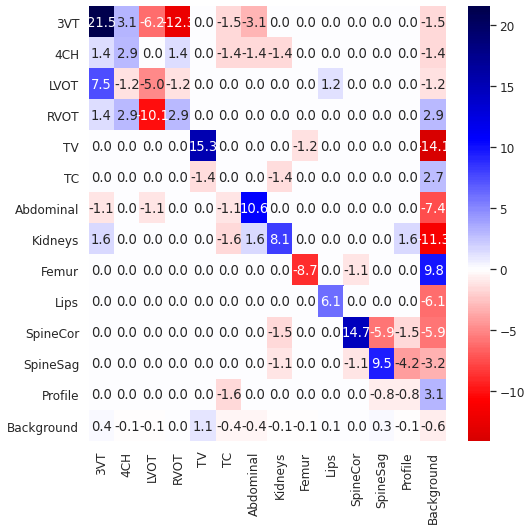}

\caption{Confusion matrix for \textit{Mix. RA} (left) and the difference in precision between \textit{Mix. RA} and \textit{SN Pol.}}
\label{fig:conf}

\end{figure}

From Table \ref{tab:results} we can see that the proposed method \textit{Mix. RA} outperforms all other augmentation methods on every metric with random network initialization, including the baseline \textit{RA} augmentation policy search. 

To better understand how \textit{Mix. RA} outperforms naive augmentation, we show the confusion matrix for the best performing model \textit{Mix. Aug} and the difference in confusion matrix between it and naive augmentation \textit{SN Pol}. We find that in general, heart plane classification is improved with a mean increase in macro F1-Score of $4.0\%$. Other anatomical planes with the exception of the femur plane also show consistent increases in performance with movement of probability mass away from erroneously classified background images to the correct classes suggesting the model is able to recognize greater variation in each anatomical class.  

\begin{figure}
\centering
\includegraphics[width=\textwidth]{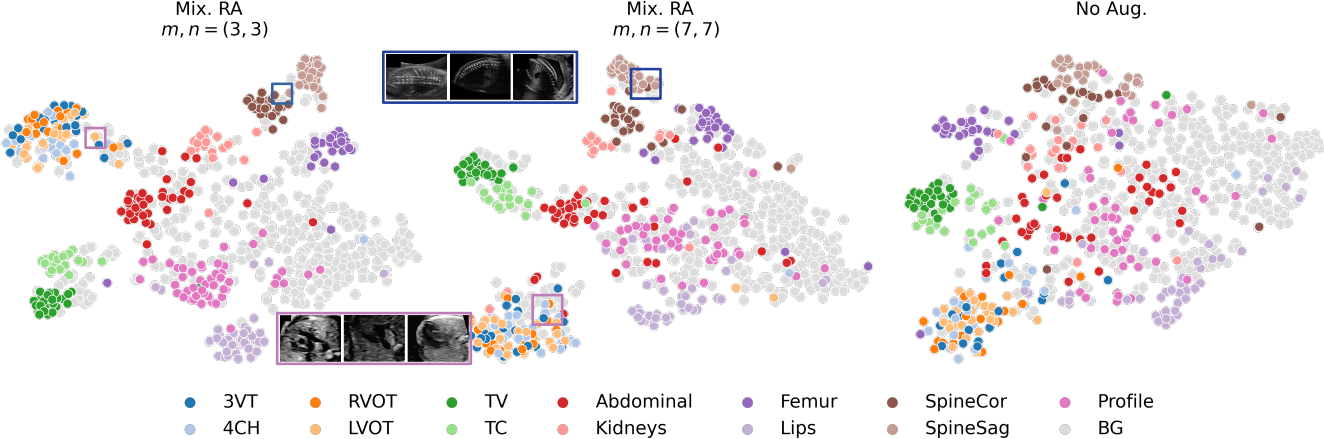}

\caption{t-SNE embeddings for different augmentation policies. The boxes represent the regions from which example images are taken from. The blue and purple boxes contain examples of the (SpineCor, SpineSag, and BG) and (3VT, RVOT, BG) classes respectively taken from the highlighted positions in the t-SNE embedding. Best viewed in color.}
\label{fig:tsne}
\end{figure}

The t-SNE embeddings of the penultimate layer seen in Figure \ref{fig:tsne} can also be used to visualize the differences in feature spaces in trained networks with different augmentation policies. Compared to the model trained with no augmentation, our best performing policy leads to a better separation of the abdominal and profile standard planes from the background class as well as clearer decision boundaries between anatomical classes. The two brain views (TC, TV) and the demarcation of the boundary between the kidney view and abdominal plane view is also better defined. 

Between the best performing policy $m, n = (5, 3)$ and an underperforming policy $m, n = (7, 7)$, we find that profile planes are better separated from the background class and the abdominal planes better separated from the kidney views, which suggests that the optimum $m, n$ value increases network recognition of salient anatomical structures. However, in all three cases, the cardiac views remain entangled. This can be attributed to the difficulty of the problem, as even human expert sonographers cannot consistently differentiate between different cardiac standard plane images. We also find that the background class also contains examples of the anatomies in each class, but in sub-optimal plane views, which leads to confusion during classification. This difficulty is illustrated in example background images in Figure \ref{fig:tsne} where the heart and spine are visible in the BG class. 

\subsubsection{Pre-trained Networks}
We also compare our work to methods where networks were initialized with external data as seen in the right of Table \ref{tab:results}. Baseline methods of self-supervised pre-training using video data \cite{jbi} \textit{(Siam. Init.)}, multi-modal saliency prediction \cite{RichardIPMI} \textit{(Saliency)} and Sononet \textit{(Sononet)} \cite{Sononet} were used to initialize the models and the models fine-tuned on our dataset. Using our augmentation policy during fine-tuning of a pre-trained SonoNet network further increased the performance of standard plane classification with an increase in final F1-score of 0.9\% when $m, n = (5, 1)$. This reduction in optimum transformation magnitude may be due to the change in network architecture from SE-ResNeXt-50 to a Sononet, as the smaller Sononet network may not be able to capture representations the former is able to. Furthermore, we find that augmentation policy with a randomly initialized network \textit{Mix. RA} approaches the performance of the \textit{Siam. Init.} and \textit{Saliency} pre-trained networks. This is despite the fact that the \textit{Siam. Init.} requires additional 135 US videos for network self-supervised initialization, and \textit{Saliency} required external multi-modal data in the form of sonographer gaze.

\subsubsection{Ablation Study}
To better understand the individual contributions to the \textit{Mix. RA} augmentation search policy, we show the results of an ablation study on the components of \textit{Mix. RA} in Table \ref{tab:ablation}. 

\begin{table}[]
\centering
\caption{Ablation study on the individual components of our \textit{Mix. RA} policy search algorithm on training of a randomly initialized CNN for ultrasound standard plane detection. All metrics are macro-averages due to the class imbalance. The \textit{Linear Mix. RA} is included as a baseline mixed-example augmentation strategy.}
\resizebox{\textwidth}{!}{%
\setlength{\tabcolsep}{4pt}
\begin{tabular}{@{}l|ccc|ccccc@{}}
\toprule
          & No Aug.      & SN Pol.      & RA           & \begin{tabular}[c]{@{}c@{}}RA + Speckle\\ \raisebox{.5pt}{\textcircled{\raisebox{-.9pt} {1}}}\end{tabular} & \begin{tabular}[c]{@{}c@{}}RA + Deform.\\ \raisebox{.5pt}{\textcircled{\raisebox{-.9pt} {2}}}\end{tabular} & \begin{tabular}[c]{@{}c@{}}Ext. RA\\ \raisebox{.5pt}{\textcircled{\raisebox{-.9pt} {1}}} $+$ \raisebox{.5pt}{\textcircled{\raisebox{-.9pt} {2}}}\end{tabular} & \begin{tabular}[c]{@{}c@{}}{\color[HTML]{666666} Linear Mix. RA}\\ {\color[HTML]{666666}\raisebox{.5pt}{\textcircled{\raisebox{-.9pt} {1}}} $+$ \raisebox{.5pt}{\textcircled{\raisebox{-.9pt} {2}}}}\end{tabular} & \begin{tabular}[c]{@{}c@{}}Non-Linear Mix. RA\\ \raisebox{.5pt}{\textcircled{\raisebox{-.9pt} {1}}} $+$ \raisebox{.5pt}{\textcircled{\raisebox{-.9pt} {2}}}\end{tabular} \\ \midrule
Precision & 56.5$\pm$1.2 & 70.4$\pm$2.3 & 73.9$\pm$1.7 & 74.6$\pm$1.7                                                  & 74.6$\pm$1.7                                                  & 74.0$\pm$2.4                                             & {\color[HTML]{666666}74.6$\pm$1.6}                                                              & \textbf{75.1$\pm$1.8}                                           \\
Recall    & 55.1$\pm$1.2 & 64.9$\pm$1.6 & 72.9$\pm$2.0 & 72.1$\pm$1.8                                                  & 72.9$\pm$1.8                                                  & \textbf{74.5$\pm$1.3 }                                            & {\color[HTML]{666666}73.2$\pm$1.5}                                                               & 73.4$\pm$1.9                                          \\
F1-Score  & 55.4$\pm$1.2 & 67.0$\pm$1.3 & 72.8$\pm$1.8 & 72.9$\pm$1.7                                                  & 73.2$\pm$1.2                                                  & 73.6$\pm$1.3                                            & {\color[HTML]{666666}73.7$\pm$1.6}                                                              & \textbf{74.0$\pm$1.8}                                           \\ \bottomrule
\end{tabular}%
}
\label{tab:ablation}
\end{table}

It can be seen that both including Speckle noise transformations and deformation (Grid, Elastic) transformations lead to increased classifier performance for standard plane classification of +0.1\% and 0.3\% respectively with further improvement when both are combined together with \textit{Ext. RA}. We find that both \textit{RA} and \textit{Ext. RA} had an optimal $m,n = (5, 3)$, suggesting that the magnitude ranges for our additional transformations are well matched to the original transformation list. This performance increase is further boosted when mixed-example augmentation is introduced on top of \textit{Ext. RA}, with non-linear mixed-example augmentations outperforming a linear mix-up based method. 

\subsubsection{Affinity and Diversity}
\begin{figure}
\centering
\includegraphics[width=\textwidth]{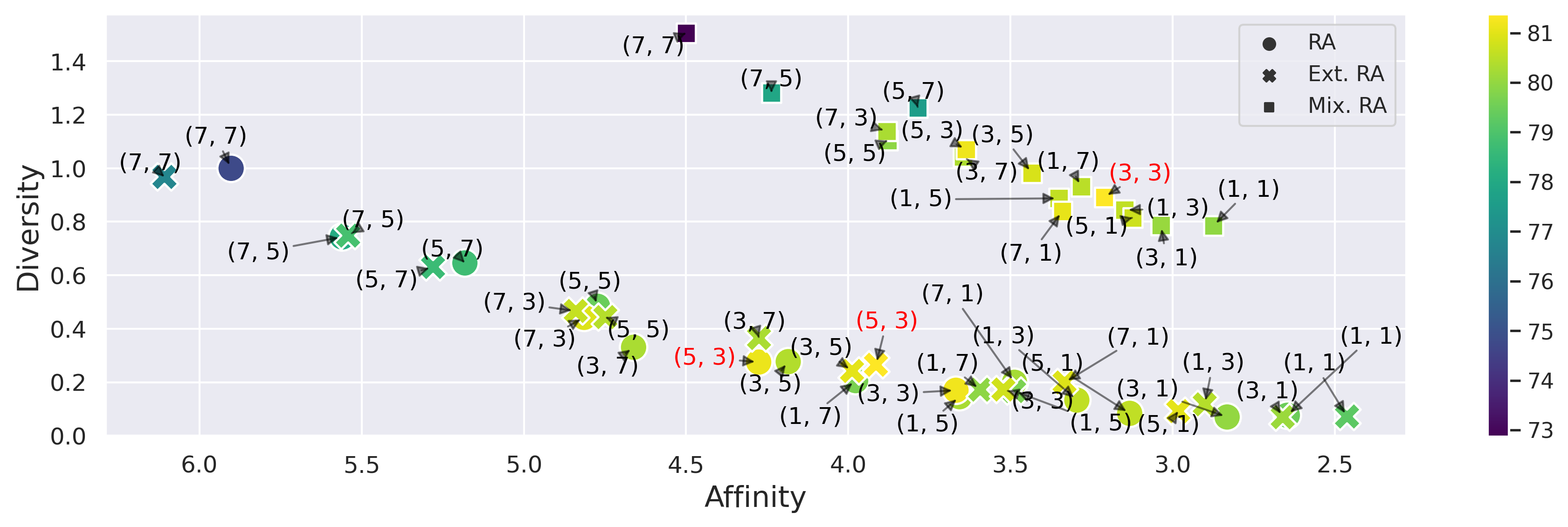}
\caption{Affinity and diversity metrics for \textit{RA}, \textit{Ext. RA} and \textit{Mix. RA} augmentation policy search algorithms. Each point represents a (m, n) value in hyperparameter space. Best performing $m, n$ values are highlighted in red for each policy search method. Color represents final F1-Score on a randomly initialized CNN.}
\label{fig:affdiv}
\end{figure}

The affinity and diversity of the augmentation policies is shown in Fig. \ref{fig:affdiv}. We find that there exists a ``sweet spot" of affinity and diversity using non-mixed class augmentation strategies at an affinity distance of $\sim$3.8 and diversity of $\sim$0.25 which maximized model performance, corresponding to $m, n = (5, 3)$. At high $m, n$ values, affinity distance is too high and the distribution of the augmented data is too far away from the validation data, decreasing model performance. However, at low $m, n$ values, the diversity of the augmented data decreases and the model sees too little variation in input data. 

It can also be seen that the \textit{Mix. RA} augmented dataset showed a reduced affinity distance to the original dataset than the \textit{Ext. RA} dataset at the same $m, n = (5, 3)$ value, implying that our proposed transforms shifts augmented images to be more similar to the original images. Moreover, using a mixed-example data augmentation strategy drastically increased diversity for any given value of data distribution affinity, which improved final model performance. The best performing mixed-example augmentation policy $m, n = (3, 3)$ reduced the magnitude of each transformation compared to the optimal non-linear augmentation policy. This suggests that mixed-example augmentation acts to increase the diversity of training images which reduces the magnitude required during further processing.

\section{Conclusion}
The results have shown that we can use a simple hyper-parameter grid search method to find an augmentation strategy that significantly outperforms conventional augmentation methods. For standard plane classification, the best performing augmentation policy had an average increase in F1-Score of 7.0\% over that of a standard ultrasound model augmentation strategy. Our augmentation policy method is competitive with the \textit{Siam. Init.} \cite{jbi} despite the latter needing additional external data for pre-training. Our method also improves the performance of a Sononet pre-trained model when fine-tuned using our augmentation policy search method. From t-SNE plots and confusion matrix differences, we can see that the performance increase is from better classification of background-labelled planes, despite the difficulty in heart plane classification. It should be noted that a large degree of misclassification was due to standard planes being mis-classified into background images or vice-versa, and qualitative evaluation of t-SNE clusters show that this was due to background images being examples of class anatomies. The  ablation study also shows that our additional transformations improve model performance, and non-linear mixed-example augmentation further improves classification. 
The evaluation using affinity and diversity indicate that the hyperparameter search involves a trade-off between diversity and affinity. We find that using non-linear mixed-class data augmentation before transformations drastically increases diversity without further increasing affinity distance between the training data and augmented data, which helps explain the increase in model performance. 
In conclusion, we have shown that our augmentation policy search method outperforms standard manual choice of augmentation policy. The augmentation policy search method presented does not have any inference-time computational cost, and has the potential to be applied in other medical image settings where training data is insufficient and costly to acquire.

\bibliographystyle{splncs04}
\bibliography{bib}

\end{document}